\documentclass[]{ceurart}

\usepackage{booktabs}
\usepackage{graphicx}
\usepackage{amsmath}
\usepackage{amssymb}
\usepackage{xspace}
\usepackage{multirow}

\begin{document}

\copyrightyear{2026}
\copyrightclause{Copyright for this paper by its authors.
  Use permitted under Creative Commons License Attribution 4.0
  International (CC BY 4.0).}

\conference{CLEF 2026 Working Notes, 21 -- 24 September 2026, Jena, Germany}

\title{Lightweight Person--Place Relation Extraction from Historical Newspapers with Dependency Graphs and Proximity Features}

\title[mode=sub]{Working Notes paper for the HIPE-2026 Shared Task at CLEF 2026}

\author[1]{Mlen-Too Wesley}[%
orcid=0009-0002-7979-1599,
email=mwesley32@gatech.edu,
]
\address[1]{Georgia Institute of Technology, Atlanta, GA, USA}

\begin{abstract}
The HIPE-2026 shared task introduces person--place relation extraction from multilingual historical newspapers as a new evaluation track, classifying the \texttt{at} and \texttt{isAt} relations between pre-annotated person and location mentions in English, French, and German.
Motivated by the cost of processing historical archives at scale, our team (DS@GT HIPE, team~2 in the official results) investigates how far a lightweight, interpretable system can go without any pretrained language model at the relation classification stage.
Our approach builds a document-level graph from dependency parses, extracts proximity-based and part-of-speech features for each entity pair, and classifies them with small scikit-learn ensembles or compact Graph Attention Networks, keeping every submitted run under 847K parameters.
On the official evaluation (Test~A, the newspaper test set), our best run reached a macro recall of 0.5142, ranking 3rd on the Efficiency profile while placing mid-table on Accuracy among the 17 participating teams.
Two findings stand out. First, minimum character distance alone captures most of the classification signal; adding further engineered features yields inconsistent gains and sometimes degrades performance, echoing prior evidence that argument distance dominates relation extraction. Second, document-grouped cross-validation is essential on this corpus: pair-level splits inflate scores by 25--37 percentage points because entity mentions recur across documents, a data-leakage effect that grouped cross-validation removes.
\end{abstract}


\maketitle

\section{Introduction}

HIPE-2026 is the third edition of the HIPE shared task series~\cite{ehrmann_extended_2020,ehrmann_extended_2022}, organized as part of CLEF~2026.
While earlier editions focused on named entity recognition (NER) and entity linking in historical documents, HIPE-2026 introduces relation extraction (RE) between pre-annotated person and location entity mentions~\cite{opitz_extended_2026,opitz_overview_2026}.\footnote{Task website: \url{https://hipe-eval.github.io/HIPE-2026/}.}
For each (person, location) pair in a document, participating systems classify two relations.
The \texttt{at} relation captures whether the text supports the inference that the person was present at the location at some point in time; it is annotated as a graded judgement with the labels TRUE, PROBABLE, or FALSE.
The \texttt{isAt} relation captures the narrower notion of the person being described as physically located at the place, annotated as a binary TRUE or FALSE.
The official evaluation metric is macro recall, also called balanced accuracy~\cite{opitz_metrics2024}, and systems are assessed under three profiles: Accuracy, Efficiency, and Generalization.
A total of 17 teams submitted 45 runs.

We approach the task with a focus on efficiency.
Historical newspaper processing at scale requires lightweight models that can run on modest hardware~\cite{schwartz_greenai2020,strubell_energy2019}, yet most relation extraction systems rely on large pretrained language models with substantial parameter budgets and inference costs.
We ask how far engineered features derived from syntactic structure can carry relation classification without any pretrained language model at the relation classification stage.
FastText embeddings~\cite{bojanowski_fasttext2017} are used during graph construction for cross-sentence bridging, but no embedding or transformer representation enters the classification pipeline itself.

\begin{figure}[t]
  \centering
  \includegraphics[width=\linewidth]{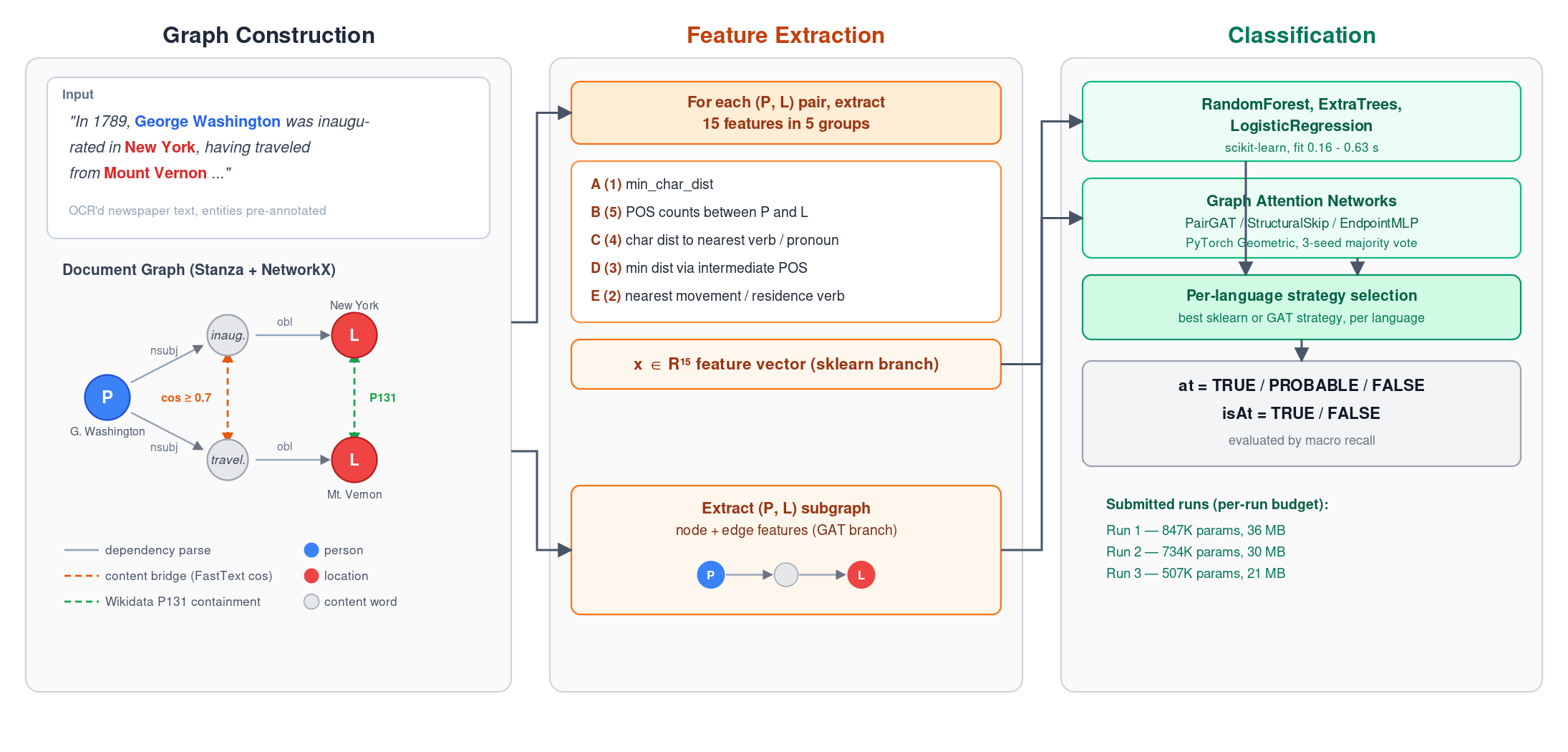}
  \caption{Three-stage pipeline. The first stage assembles Stanza dependency parses into a document-level graph with cross-sentence bridges (content-word cosine similarity and Wikidata P131 containment). The second stage extracts 15 proximity-based features and a per-pair subgraph for each entity pair. The third stage classifies via scikit-learn ensembles (from the feature vector) or Graph Attention Networks (from the subgraph), with per-language strategy selection.}
  \label{fig:pipeline}
\end{figure}

Figure~\ref{fig:pipeline} illustrates the system architecture.
Our system follows a three-stage pipeline consisting of graph construction from dependency parses, extraction of 15 proximity-based and POS-based features, and classification via scikit-learn ensembles~\cite{pedregosa_sklearn2011} or small Graph Attention Networks built with PyTorch Geometric~\cite{fey_pyg2019}.\footnote{Code: \url{https://github.com/dsgt-arc/hipe-2026}}
Per-run parameter counts range from 506,923 to 846,701, with model sizes of 21--36~MB.
Run~1 achieved a macro recall of 0.5142, ranking 26th of 46 on accuracy and 3rd overall in the efficiency profile.

The remainder of this paper is organized as follows.
Section~2 surveys related work.
Section~3 describes the data.
Section~4 presents the system in detail.
Section~5 reports experiments and results.
Section~6 concludes.

\section{Related Work}

\paragraph{HIPE shared task lineage.}
CLEF HIPE-2020 introduced NER evaluation on historical newspapers with 13 participating teams~\cite{ehrmann_extended_2020}, and HIPE-2022 expanded to NER and entity linking across multiple multilingual historical datasets~\cite{ehrmann_extended_2022}.
HIPE-2026~\cite{opitz_extended_2026,opitz_overview_2026} introduces relation extraction as a new task type.
OCR noise, domain heterogeneity, and diachronic language change remain persistent challenges for NLP on historical text~\cite{ehrmann_ner_survey2023,hamdi_ocr_impact2023}, and these challenges compound for RE, where upstream parse errors propagate into both graph structure and feature quality.

\paragraph{Dependency-based and graph-based RE.}
Bunescu and Mooney~\cite{bunescu_sdp2005} established that the shortest dependency path between two entities carries strong signal for relation classification.
Subsequent work applied graph neural networks over dependency trees. Zhang et al.~\cite{zhang_cgcn2018} proposed C-GCN over pruned trees, and Guo et al.~\cite{guo_aggcn2019} introduced attention-guided graph convolutions.
Nan et al.~\cite{nan_latent2020} constructed document-level graphs from dependency parses using spaCy and NetworkX, the closest published precedent to our graph construction approach.
Peng et al.~\cite{peng_crosssentence2017} addressed cross-sentence RE through graph LSTMs over merged sentence-level dependencies. Document-level RE benchmarks such as DocRED~\cite{yao_docred2019} formalize the multi-sentence setting that our document graphs target.
Our system differs from these approaches in that it uses the graph primarily for feature computation rather than learned message passing, with the GAT branch as a secondary pathway.

\paragraph{Feature engineering and proximity dominance.}
Jiang and Zhai~\cite{jiang_features2007} systematically explored RE feature spaces and found that complex subspaces provide marginal gains over simpler sets; Zhou et al.~\cite{zhou_features2005} catalogued the relative contributions of lexical, chunk, and parse-tree features.
Zeng et al.~\cite{zeng_position2014} showed that position features are the single most informative group for relation classification, position-aware attention over the TACRED benchmark likewise relies heavily on argument position~\cite{zhang_tacred2017}, and Alt et al.~\cite{alt_probing2020} confirmed via probing experiments that argument distance accounts for more RE performance than verb or POS semantics.
These findings align with our result that minimum character distance alone captures most of the classification signal (Section~\ref{sec:ablation}).
For biographical RE specifically, de-Dios-Flores et al.~\cite{de_dios_flores_biographical2015} found that basic lemmatized features outperform deep syntactic dependency features for person-place relations in Portuguese and Spanish, which directly parallels our negative finding on verb semantic features (Group~E).

\paragraph{Lightweight RE and historical/biographical text.}
Giuliano et al.~\cite{giuliano_shallow2006} showed that shallow linguistic features consistently outperform deep syntactic parsing for RE on biomedical text, where parser errors propagate through complex feature representations; OCR-degraded historical text presents an analogous noise profile.
Minard et al.~\cite{minard_svm2011} achieved 3rd place in the i2b2 2010 clinical RE shared task using SVMs over engineered surface and distance features with no neural components. Lightweight feature-based systems such as LightRel~\cite{roth_lightrel2018} likewise remain competitive on standard relation-classification benchmarks with minimal engineering.
Plum~\cite{plum_biographical2022} developed language-agnostic biographical RE using Wikipedia-to-Wikidata alignment, providing precedent for our Wikidata containment edges.
HIPE-2026 is, to our knowledge, the first shared task targeting person--place RE from historical newspapers, and its explicit efficiency evaluation profile~\cite{schwartz_greenai2020,strubell_energy2019} aligns with this lightweight tradition.

\section{Data}

The HIPE-2026 training data is drawn from the Impresso collection of digitized historical newspapers spanning the 19th and 20th centuries~\cite{opitz_extended_2026,opitz_overview_2026}.
Person and location entity mentions are pre-annotated. The task is to classify the relation between sampled (person, location) pairs, not to identify the entities.
Table~\ref{tab:data} summarizes the training data.

\begin{table}[t]
  \caption{Training data by language, showing document and entity pair counts from the Impresso corpus.}
  \label{tab:data}
  \begin{tabular}{lrr}
    \toprule
    Language & Documents & Entity Pairs \\
    \midrule
    English  & 35 & 307 \\
    French   & 35 & 478 \\
    German   & 34 & 466 \\
    \midrule
    Total    & 104 & 1,251 \\
    \bottomrule
  \end{tabular}
\end{table}

The test data consists of Impresso newspaper test sets for each language, with sizes determined by the organizers (collectively, Test~A, the newspaper test set), plus a surprise French literary test set (Test~B).\footnote{We follow the organizers' dataset naming throughout: Test~A / Newspaper test set and Test~B / French literary surprise test set.}
We addressed the surprise set by reusing models trained on French newspaper data without domain adaptation.

Three characteristics of the corpus shaped our system design.
First, OCR noise degrades entity surface forms and surrounding context~\cite{ehrmann_ner_survey2023,hamdi_ocr_impact2023}.
Second, the training sets are small, with only hundreds of entity pairs per language, which limits the viability of data-hungry approaches.
Third, entity mentions recur across documents within the same newspaper, creating the potential for information leakage in naive cross-validation splits (Section~\ref{sec:leakage}).

\section{System Description}

\subsection{Graph Construction}
\label{sec:graph}

The system constructs a document-level graph for each input document as follows.
First, each sentence is parsed with Stanza~\cite{qi_stanza2020}, a neural NLP toolkit providing tokenization, part-of-speech tagging, and dependency parsing, using its Universal Dependencies models for English, French, and German. Each sentence becomes a directed tree in a NetworkX DiGraph.
We map multi-token entity mention spans to their syntactic head node.

Three bridging mechanisms connect otherwise disjoint sentence-level trees into a single document graph.
\emph{Entity merging} collapses nodes that share the same pre-annotated entity identifier, that is, different mentions of the same person or location, into a single node.
\emph{Content-word linking} identifies non-stopword content words (tagged NOUN or PROPN) from different sentences whose FastText~\cite{bojanowski_fasttext2017} embeddings have a cosine similarity $\geq 0.7$ and merges them, creating cross-sentence edges.
\emph{Geographic containment} adds edges between location entities that stand in a Wikidata P131 (``located in the administrative territorial entity'') relationship.

Table~\ref{tab:bridges} shows the effect of each bridging mechanism.
Content-word linking is the dominant mechanism. Removing it increases the proportion of disconnected entity pairs from 30.6\% to 64.5\%.
Without any bridging, 77.0\% of entity pairs are in disconnected graph components.

\begin{table}[t]
  \caption{Bridging ablation. Each row removes one mechanism from the full pipeline. ``Disconnected pairs'' is the percentage of entity pairs with no graph path between them.}
  \label{tab:bridges}
  \begin{tabular}{lrr}
    \toprule
    Configuration & Disconnected (\%) & Avg.\ components \\
    \midrule
    Full pipeline              & 30.6 & 4.7 \\
    Skip entity merging        & 32.1 & 5.0 \\
    Skip content-word linking  & 64.5 & 11.8 \\
    Skip Wikidata containment  & 36.1 & 4.9 \\
    Skip all bridges           & 77.0 & 14.2 \\
    \bottomrule
  \end{tabular}
\end{table}

\subsection{Feature Extraction}
\label{sec:features}

From each document graph, we extract 15 features per entity pair, organized into five groups (Table~\ref{tab:features}).
All features are proximity-based or POS-based; no embedding features or transformer representations enter the classification pipeline.
The feature design draws on prior work showing that distance-based features dominate in relation classification~\cite{zhou_features2005,jiang_features2007,zeng_position2014}.

\begin{table}[t]
  \caption{Feature groups A--E. All 15 features are computed from the document graph and surface-level token positions.}
  \label{tab:features}
  \begin{tabular}{clp{5.5cm}}
    \toprule
    Group & \# & Description \\
    \midrule
    A & 1 & Minimum character distance between any mention pair of the two entities \\
    B & 5 & POS-tagged token counts between entities (verbs, nouns, proper nouns, pronouns) and a binary pronoun indicator \\
    C & 4 & Character distances from each entity to the nearest verb and pronoun \\
    D & 3 & Minimum character distances routed through intermediary verbs, nouns, and pronouns \\
    E & 2 & Character distance to the nearest movement and residence verb \\
    \bottomrule
  \end{tabular}
\end{table}

\subsection{Scikit-learn Classifiers}
\label{sec:sklearn}

Each scikit-learn strategy trains two relation classifiers on the 15 features, one for \texttt{at} (3-class) and one for \texttt{isAt} (binary).
We use three classifier types, all implemented with scikit-learn~\cite{pedregosa_sklearn2011}.
RandomForest uses 500 estimators with max depth~10; ExtraTrees uses 300 estimators with max depth~10 or unbounded; LogisticRegression uses L2 penalty with $C{=}1.0$ and StandardScaler preprocessing.
All classifiers use balanced class weights (\texttt{class\_weight="balanced"}) to handle label imbalance.

Training variants include per-language models and a multilingual-concatenated variant (pass\_c) that trains on data from all three languages.
Training times range from 0.16 to 0.63 seconds per strategy.

\subsection{Graph Attention Networks}
\label{sec:gat}

We evaluated four architectures based on the Graph Attention Network~\cite{velickovic_gat2018}, implemented with PyTorch Geometric~\cite{fey_pyg2019}, all operating on per-pair subgraphs extracted from the document graph.
Each node carries a 336-dimensional feature vector (300 from FastText plus 36 structural one-hots encoding node kind, pair role, alignment status, and Universal Dependencies POS tag).
Each edge carries a 38-dimensional feature vector encoding the dependency relation label, direction, and edge type.
All architectures jointly predict both \texttt{at} and \texttt{isAt} through a shared hidden layer that feeds separate classification heads.

\paragraph{PairGAT.}
The base architecture projects the 336-dim node features through a linear layer into a hidden space of $\text{hidden\_dim} \times \text{n\_heads}$ dimensions (default: $32 \times 4 = 128$), then applies two GATv2Conv layers~\cite{brody_gatv2_2022} with LayerNorm and dropout after each.
GATv2Conv was chosen over the original GATConv because the v2 variant fixes a known limitation in which all attention heads attend to the same neighbor regardless of query.
The person and location endpoint embeddings are concatenated into a pair representation, optionally augmented by a scalar side-channel that projects proximity features through a small feed-forward network.
This side-channel handles the 30\% of entity pairs with no graph path. When the subgraph is degenerate, the GAT produces near-zero embeddings and the side-channel carries the prediction.
A shared linear layer with ReLU and dropout (0.5) feeds two output heads for \texttt{at} (3-class) and \texttt{isAt} (binary).

\paragraph{PairGATBranched.}
In PairGAT, the FastText block occupies 300 of 336 input dimensions (89\%), so the structural one-hots compete for only 11\% of the input projection capacity.
PairGATBranched addresses this by projecting the FastText and structural slices through separate linear layers (64-dim and 32-dim respectively) before concatenation, giving structural features a fixed 33\% share of the hidden space.
The concatenated 96-dim representation then passes through the same GATv2Conv layers and readout as PairGAT.

\paragraph{PairGATStructuralSkip.}
This variant uses the same GAT backbone as PairGAT but concatenates the raw structural features of the person and location endpoints (a skip connection of $2 \times 36 = 72$ dimensions) alongside the GAT-produced embeddings at the readout layer.
The motivation is to test whether structural information (node kind, role, POS) is lost during message passing.
If the skip connection helps, the GAT layers are diluting information that matters for classification.
This variant is less invasive than PairGATBranched because the GAT layers themselves are unchanged.

\paragraph{EndpointMLP.}
A non-graph baseline that concatenates the raw 336-dim feature vectors of the person and location endpoints ($2 \times 336 = 672$ dimensions) and passes them through a feed-forward network with no message passing or attention.
If EndpointMLP matches the GAT architectures, it indicates that graph reasoning adds no value beyond what the endpoint features already encode.
That the non-graph EndpointMLP was selected for French by the newspaper held-out sweep (Table~\ref{tab:gat_sweep}) and used in the submitted French GAT run suggests that for some language/corpus combinations the graph topology carries little additional signal.

We selected architectures per language via document-grouped 5-fold cross-validation (StratifiedGroupKFold), following a similar small-data setup to Mandya et al.~\cite{mandya_gat2020}.
Table~\ref{tab:gat_cv} shows the selected architecture per language.

\begin{table}[t]
  \caption{GAT architecture selection via document-grouped 5-fold cross-validation on the training data; we report the CV winner per language. The submitted runs used the architecture chosen by the newspaper held-out sweep (Table~\ref{tab:gat_sweep}), which matches the CV winner for English but differs for French and German (see Section~\ref{sec:runs}).}
  \label{tab:gat_cv}
  \begin{tabular}{llrl}
    \toprule
    Language & Winner & CV macro avg & Wall time (s) \\
    \midrule
    EN & PairGATStructuralSkip & $0.549 \pm 0.047$ & 53 \\
    FR & PairGAT              & $0.557 \pm 0.018$ & 599 \\
    DE & EndpointMLP           & $0.544 \pm 0.022$ & 35 \\
    \bottomrule
  \end{tabular}
\end{table}

All GAT variants use hidden\_dim\,=\,32, n\_heads\,=\,4, n\_layers\,=\,2, dropout\,=\,0.3, attention\_dropout\,=\,0.3, and head\_dropout\,=\,0.5.
PairGATBranched uses fasttext\_hidden\,=\,64 and structural\_hidden\,=\,32.
All architectures share a 128-dim shared head layer before the classification outputs.
We produce GAT predictions by 3-seed majority vote to reduce variance from random initialization.
Parameter counts range from 335,247 (EndpointMLP) to 511,503 (PairGATStructuralSkip), with model sizes of 5--8~MB per architecture.

\subsection{Strategy Portfolio and Run Composition}
\label{sec:runs}

Each of the three submitted runs combines one strategy per language.
Table~\ref{tab:runs} shows the mapping, and Table~\ref{tab:budgets} shows the resource budget per run.
All three runs reuse the corresponding French strategy for the surprise literary French test set.

\begin{table}[t]
  \caption{Run-to-strategy mapping. Each cell names the strategy used for that language in that run.}
  \label{tab:runs}
  \begin{tabular}{llll}
    \toprule
    & Run 1 & Run 2 & Run 3 \\
    \midrule
    EN & GAT StructuralSkip  & RF + ExtraTrees & ExtraTrees + LR \\
    FR & LR + ExtraTrees      & RF + ExtraTrees & GAT EndpointMLP \\
    DE & RF + ET (multilingual) & GAT PairGAT   & RF + ExtraTrees \\
    \bottomrule
  \end{tabular}
\end{table}

\begin{table}[t]
  \caption{Per-run resource budgets. Parameter counts are summed from the strategies used in each run; surprise-FR reuses the FR model and is not double-counted.}
  \label{tab:budgets}
  \begin{tabular}{lrr}
    \toprule
    Run & Parameters & Model size (MB) \\
    \midrule
    Run 1 & 846,701 & 36 \\
    Run 2 & 733,751 & 30 \\
    Run 3 & 506,923 & 21 \\
    \bottomrule
  \end{tabular}
\end{table}

\section{Experiments and Results}

\subsection{Evaluation Setup}
\label{sec:evalsetup}
The evaluation metric is macro recall, also called balanced accuracy; the per-relation scores are $\mathrm{MacroRecall}_{\mathtt{at}}$ and $\mathrm{MacroRecall}_{\mathtt{isAt}}$, and we report a per-language \emph{global score} equal to their arithmetic mean. The overall Test~A score averages the global score across English, French, and German.
We report results under three protocols, and each results-table caption states which one it uses: (i)~\emph{document-grouped 5-fold cross-validation} (StratifiedGroupKFold) on the training data, used for model selection and ablation; (ii)~an \emph{internal newspaper held-out set} of 7 documents per language, used for cross-strategy comparison before submission; and (iii)~the \emph{official evaluation} on Test~A (newspaper) and Test~B (French literary surprise set).
Our scikit-learn strategies come in three passes: \emph{pass~A} trains a fixed classifier pair per language, \emph{pass~B} selects the per-language cross-validation winner for each relation, and \emph{pass~C} trains on multilingual-concatenated data from all three languages.

\subsection{Cross-Validation Protocol and Leakage Finding}
\label{sec:leakage}

Our initial cross-validation protocol used StratifiedKFold with pair-level splits.
We discovered that entity mentions recur across documents within the same newspaper, so a person or location appearing in one training document often appears in a held-out document as well.
Naive CV splits that ignore this structure leak document-specific patterns into the validation set, producing inflated performance estimates~\cite{elangovan_leakage2021}.

\begin{figure}[t]
  \centering
  \includegraphics[width=0.82\linewidth]{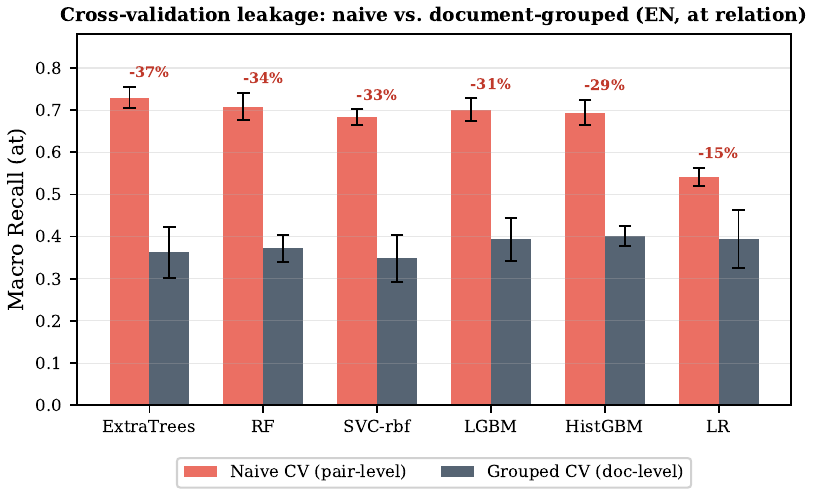}
  \caption{Naive pair-level CV versus document-grouped CV for four classifier/relation combinations on English data. The gap ranges from 15 to 37 percentage points.}
  \label{fig:leakage}
\end{figure}

We corrected this by switching to StratifiedGroupKFold, grouping by document, with 5 folds~\cite{roberts_cv2017,sogaard_splits2021}.
Table~\ref{tab:leakage} shows the magnitude of the inflation on English data.
The gap ranges from 15 to 37 percentage points depending on the classifier and relation.
Tree-based models suffer the largest drops (ExtraTrees loses 37\,pp on \texttt{at}), because they can memorize document-specific feature patterns when the same document's entity pairs appear in both training and validation folds.
LogisticRegression, which cannot capture such document-level structure as easily, drops only 15\,pp.
This differential confirms that document-grouped evaluation is essential for this task and that the magnitude of inflation depends on the model family.
Figure~\ref{fig:leakage} visualizes the gap across four classifier/relation combinations.

\begin{table}[t]
  \caption{CV leakage on English data. Naive CV uses pair-level StratifiedKFold; grouped CV uses StratifiedGroupKFold by document. Gap is in percentage points.}
  \label{tab:leakage}
  \begin{tabular}{llrrr}
    \toprule
    Classifier & Relation & Naive CV & Grouped CV & Gap \\
    \midrule
    ExtraTrees & \texttt{at}   & 0.730 & 0.363 & $-$37 \\
    RF         & \texttt{at}   & 0.709 & 0.372 & $-$34 \\
    ExtraTrees & \texttt{isAt} & 0.807 & 0.527 & $-$28 \\
    LR         & \texttt{at}   & 0.542 & 0.395 & $-$15 \\
    \bottomrule
  \end{tabular}
\end{table}

\subsection{Feature Ablation}
\label{sec:ablation}

We evaluated feature group contributions using HistGBM (500 iterations, learning rate 0.05) with balanced sample weights and document-grouped 5-fold CV.
Table~\ref{tab:ablation} shows the cumulative tier ablation, where each column adds one more feature group to the classification pipeline.

\begin{table*}[t]
  \caption{Cumulative feature ablation. Each column adds one more feature group. Values are mean macro recall $\pm$ standard deviation across 5 folds of document-grouped CV.}
  \label{tab:ablation}
  \begin{tabular}{llrrrrr}
    \toprule
    Lang & Rel & A & A+B & A+B+C & A+B+C+D & Full (A--E) \\
    \midrule
    EN & \texttt{at}   & $.325 \pm .042$ & $.313 \pm .047$ & $.352 \pm .043$ & $.359 \pm .062$ & $.348 \pm .059$ \\
    EN & \texttt{isAt} & $.471 \pm .057$ & $.553 \pm .046$ & $.523 \pm .034$ & $.546 \pm .060$ & $.542 \pm .084$ \\
    FR & \texttt{at}   & $.402 \pm .013$ & $.395 \pm .017$ & $.393 \pm .010$ & $.410 \pm .009$ & $.408 \pm .014$ \\
    FR & \texttt{isAt} & $.592 \pm .031$ & $.586 \pm .032$ & $.557 \pm .021$ & $.552 \pm .017$ & $.547 \pm .012$ \\
    DE & \texttt{at}   & $.408 \pm .050$ & $.422 \pm .059$ & $.401 \pm .069$ & $.397 \pm .059$ & $.385 \pm .053$ \\
    DE & \texttt{isAt} & $.530 \pm .077$ & $.557 \pm .072$ & $.513 \pm .034$ & $.511 \pm .034$ & $.504 \pm .030$ \\
    \bottomrule
  \end{tabular}
\end{table*}

Group~A alone (minimum character distance) captures most of the classification signal.
Adding further groups provides inconsistent gains across languages and relations.
For DE \texttt{at}, performance degrades from 0.408 with Group~A alone to 0.385 with the full feature set.
For FR \texttt{isAt}, it declines from 0.592 to 0.547, a drop of 4.5 percentage points.
These results align with prior findings that distance features dominate in relation classification~\cite{jiang_features2007,zeng_position2014} and that complex features provide marginal or negative returns~\cite{alt_probing2020}.
The pattern also echoes the findings of de-Dios-Flores et al.~\cite{de_dios_flores_biographical2015}, who reported that basic lemmatized features outperformed deep syntactic features for biographical person-place extraction; our Group~E verb-category features, which encode the closest analog to deep semantic knowledge in our pipeline, provide no consistent benefit.

To isolate the marginal contribution of each group, we also performed leave-one-group-out ablation. For each group, we trained the full pipeline with that group removed and computed the difference from the full model score.
Table~\ref{tab:logo} reports these deltas in percentage points. Positive values indicate that the group helped, meaning that removing it dropped performance.

\begin{table}[t]
  \caption{Leave-one-group-out marginal contributions (percentage points). Positive values mean the group helped; negative values mean it was noise or harmful. Deltas smaller than $\sim$2\,pp are within fold variance.}
  \label{tab:logo}
  \begin{tabular}{llrrrrr}
    \toprule
    Lang & Rel & $-$A & $-$B & $-$C & $-$D & $-$E \\
    \midrule
    EN & \texttt{at}   & $-$0.4 & $-$0.2 & +2.8 & $-$0.9 & $-$1.2 \\
    EN & \texttt{isAt} & +2.2 & +1.4 & $-$0.8 & +1.3 & $-$0.4 \\
    FR & \texttt{at}   & $-$0.4 & +0.7 & +0.8 & +1.9 & $-$0.1 \\
    FR & \texttt{isAt} & +0.1 & +1.1 & $-$1.2 & +0.3 & $-$0.5 \\
    DE & \texttt{at}   & +0.1 & $-$0.4 & $-$1.8 & $-$3.2 & $-$1.2 \\
    DE & \texttt{isAt} & +0.5 & +0.6 & $-$0.5 & $-$1.0 & $-$0.7 \\
    \bottomrule
  \end{tabular}
\end{table}

\begin{figure}[t]
  \centering
  \includegraphics[width=0.82\linewidth]{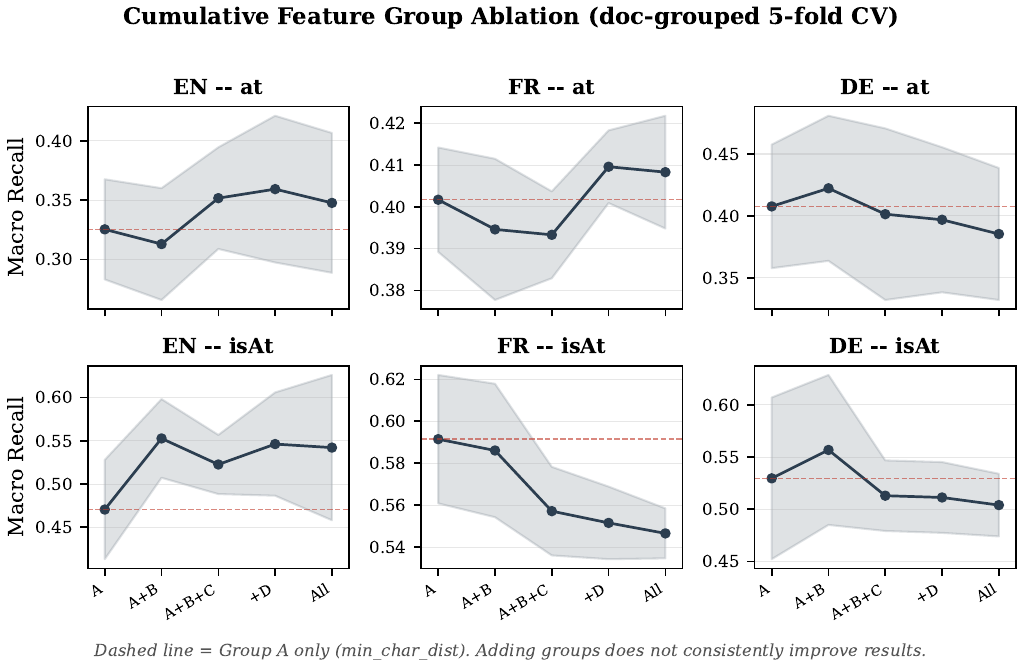}
  \caption{Cumulative feature ablation across languages and relations. Each line adds one feature group from left to right. Shaded bands show $\pm$1 standard deviation across 5 folds.}
  \label{fig:ablation}
\end{figure}

No single feature group helps uniformly across all six (language, relation) cells.
Group~A (minimum character distance) has the clearest positive signal for English \texttt{isAt} (+2.2\,pp) but is within noise for most other cells.
Group~D (intermediary-routed distances) helps French \texttt{at} (+1.9\,pp) but actively hurts German \texttt{at} ($-$3.2\,pp), the largest negative delta in the table.
Group~E (verb-category features) provides no clear benefit anywhere, consistent with its reliance on a hand-curated lexicon that may not transfer well across languages.
The DE \texttt{at} cell is the most striking example. Every group beyond A either has no effect or degrades performance, confirming the over-engineering pattern visible in the cumulative ablation (Figure~\ref{fig:ablation}).

RandomForest feature importance scores corroborate the ablation.
For \texttt{at}, the top four features by importance are \textit{l\_position\_norm} (0.151), \textit{min\_char\_dist} (0.150), \textit{p\_position\_norm} (0.124), and \textit{sent\_dist} (0.089); all four are proximity-based, and together they account for over 50\% of total importance.
For \texttt{isAt}, the same four features dominate, with \textit{min\_char\_dist\_via\_pron} (0.051) entering the top five.
The pronoun-bridge feature ranking is notable. It confirms that for \texttt{isAt}, the path from a person entity through a pronoun to a location carries discriminative signal, consistent with the observation that physical presence is often expressed through pronominal co-reference rather than direct mention.
By contrast, features from Group~E (verb-category distances) rank near the bottom for both relations, explaining why their removal in the leave-one-group-out analysis produces negligible or negative deltas.

\subsection{Official Results}
\label{sec:results}

Our best submission, Run~1, reached a macro recall of 0.5142 on Test~A, placing our team (DS@GT HIPE, team~2) 26th of the 46 ranked entries on the Accuracy profile and 3rd on the Efficiency profile; the 46 entries comprise 45 participant runs from 17 teams plus the organizers' baseline. We refer the reader to the overview papers for the full leaderboard~\cite{opitz_extended_2026,opitz_overview_2026}.
Table~\ref{tab:accuracy} presents the Accuracy profile results.
All three runs score above the random baseline (0.4049) but below the Mistral-3B baseline (0.5818)~\cite{opitz_metrics2024}.
The top-performing system achieved 0.7479.

\begin{table}[t]
  \caption{Accuracy profile results on Test~A (newspaper test set); official evaluation.}
  \label{tab:accuracy}
  \begin{tabular}{lrr}
    \toprule
    Run & Macro Recall & Rank (of 46) \\
    \midrule
    Run 1 & 0.5142 & 26 \\
    Run 2 & 0.4836 & 30 \\
    Run 3 & 0.4771 & 31 \\
    \bottomrule
  \end{tabular}
\end{table}

Table~\ref{tab:perlang} breaks down the accuracy results by language and relation.
Two patterns emerge.
First, \texttt{isAt} is consistently easier than \texttt{at} across all languages and runs, with scores 0.10--0.24 points higher.
The binary \texttt{isAt} distinction (physically present or not) is more directly expressed by proximity features than the three-way \texttt{at} judgment (which requires abductive reasoning about whether a person has \emph{ever} been at a location).
Second, no single run dominates across all languages. Run~1 produces the best English \texttt{isAt} score (0.622) but its French \texttt{at} (0.397) trails Run~2's (0.415), and Run~3 is weakest overall on newspaper data but produces the best surprise French score (0.392).
The run ordering reversal on Test~B suggests that the GAT EndpointMLP strategy used by Run~3 for French generalizes better to out-of-domain literary text than the sklearn strategies in Runs~1 and~2, consistent with the node-level FastText representation capturing some domain-transferable signal even without graph message passing.

\begin{table*}[t]
  \caption{Per-language accuracy breakdown showing \texttt{at} and \texttt{isAt} macro recall components and per-language rank (of 46 runs). Columns EN/FR/DE are Test~A; ``Surp.\ FR'' is Test~B. Official evaluation; Surprise FR reuses the corresponding FR strategy.}
  \label{tab:perlang}
  \begin{tabular}{lrrrrrrrrrr}
    \toprule
    & \multicolumn{3}{c}{English} & \multicolumn{3}{c}{French} & \multicolumn{3}{c}{German} & Surp.\ FR \\
    \cmidrule(lr){2-4} \cmidrule(lr){5-7} \cmidrule(lr){8-10} \cmidrule(lr){11-11}
    Run & \texttt{at} & \texttt{isAt} & Rank & \texttt{at} & \texttt{isAt} & Rank & \texttt{at} & \texttt{isAt} & Rank & Avg \\
    \midrule
    Run 1 & .399 & .622 & 27 & .397 & .635 & 26 & .428 & .604 & 28 & .363 \\
    Run 2 & .371 & .510 & 40 & .415 & .619 & 25 & .390 & .598 & 33 & .372 \\
    Run 3 & .363 & .604 & 30 & .414 & .471 & 37 & .420 & .590 & 30 & .392 \\
    \bottomrule
  \end{tabular}
\end{table*}

Table~\ref{tab:efficiency} shows the efficiency profile.
Run~1 ranked 3rd overall, with per-language efficiency ranks of 2nd in English and 5th in German; Run~2 ranked 2nd in French.
The efficiency ranking combines accuracy rank, parameter count rank, and model size rank with equal weight.
The parameter counts reported to the organizers used a global total (2,087,375 parameters, 87~MB) identical across all three runs rather than the per-run values in Table~\ref{tab:budgets}. This global figure, not the per-run counts, is what produced our official Efficiency ranks; we therefore present the per-run budgets of 507K to 847K parameters as a corrected re-analysis. Because the Efficiency profile is a relative ranking against other teams, we cannot recompute our exact rank from the corrected figures, but since our true per-run footprint is smaller than the submitted 2.09M total, our genuine efficiency is at least as good as the reported rank.

\begin{table}[t]
  \caption{Efficiency profile results on Test~A (official evaluation). Ranks are among the 46 ranked entries. Parameter and size ranks reflect global totals reported to organizers rather than per-run values.}
  \label{tab:efficiency}
  \begin{tabular}{lrrrr}
    \toprule
    Run & Eff.\ Rank & Acc.\ Rank & Param Rank & Size Rank \\
    \midrule
    Run 1 & \textbf{3} & 23 & 5 & 4 \\
    Run 2 & 4 & 27 & 5 & 4 \\
    Run 3 & 5 & 28 & 5 & 4 \\
    \bottomrule
  \end{tabular}
\end{table}

On Test~B (the French literary surprise set), all three runs score below 0.40 (Table~\ref{tab:perlang}, last column), reflecting the domain shift from 19th--20th century newspapers to 18th century literary text.
Run~3 (GAT EndpointMLP, 0.392) outperforms Run~1 (sklearn, 0.363), reversing the newspaper ordering.
The EndpointMLP architecture operates on raw node feature vectors without graph message passing, which may reduce its dependence on newspaper-specific parse patterns. The absolute Test~B scores nonetheless remain low for all runs, and the failure modes differ from the newspaper setting: the literary register uses longer, more subordinated sentences and archaic orthography that degrade the Stanza parses further, so a larger share of entity pairs fall in disconnected graph components, and the \texttt{at} label distribution shifts away from the newspaper prior the classifiers were tuned on. A dedicated domain-adaptation step, rather than reusing the French newspaper models unchanged, is the most promising remedy.

\subsection{Internal Model Evaluation}
\label{sec:scorecard}

Before submission, we evaluated all strategies on a newspaper-only held-out set (7 documents per language, with the remaining 27--28 documents used for training) using the official HIPE scorer.
Table~\ref{tab:scorecard} summarizes results across four model tiers.

\begin{table}[t]
  \caption{Internal evaluation scorecard (newspaper held-out, 7 docs/lang). Mean macro recall across EN, FR, DE. Tiers reflect increasing methodological complexity.}
  \label{tab:scorecard}
  \begin{tabular}{llr}
    \toprule
    Tier & Approach & Mean \\
    \midrule
    \multirow{2}{*}{1: Baseline}
      & Heuristic          & 0.442 \\
      & Majority class      & 0.417 \\
    \midrule
    \multirow{2}{*}{2: Rule}
      & Path length tiered  & 0.549 \\
      & Same subtree        & 0.462 \\
    \midrule
    \multirow{3}{*}{3: Tabular}
      & sklearn pass\_C (multilingual) & 0.520 \\
      & sklearn pass\_A (per-lang)     & 0.514 \\
      & sklearn pass\_B (per-lang)     & 0.507 \\
    \midrule
    3: Graph & GAT 3-seed majority & 0.519 \\
    \bottomrule
  \end{tabular}
\end{table}

The most consequential finding from internal evaluation is that the path-length-tiered rule (Tier~2) achieves 0.549, outperforming all trained models on mean score.
This rule classifies pairs as TRUE when the graph path between them is short and FALSE when long, using empirically tuned thresholds.
That a deterministic rule with no trainable parameters beats both sklearn ensembles and GATs is consistent with the feature ablation finding. Minimum character distance carries most of the classification signal, and the path-length rule is essentially a graph-topology encoding of the same proximity principle.
The 15 additional features and the learned decision boundaries of the trained models do not overcome the noise they introduce.

The trained models (Tier~3) cluster tightly between 0.507 and 0.520, with the GAT 3-seed majority vote (0.519) essentially matching the best sklearn pass (0.520).
Per-language results diverge in a pattern consistent with the parse quality analysis (Section~\ref{sec:parse}). The best English strategy is path\_length\_tiered (0.570, Tier~2), exploiting the fact that when English parses are noisy, a simple rule over the graph topology outperforms learned models.
French and German are best served by sklearn pass\_B (0.583) and sklearn pass\_C (0.519) respectively, where cleaner parses and more training data allow learned models to surpass the rule.
This per-language variation motivated our run composition strategy of selecting independently per language rather than using a single global model.

\subsection{Efficiency Analysis}
\label{sec:efficiency}

The classification step uses no pretrained language model.
Stanza~\cite{qi_stanza2020} dependency parsing is the only neural component in the inference pipeline, and its output serves as input to deterministic graph construction and feature extraction rather than providing learned representations for classification.
FastText~\cite{bojanowski_fasttext2017} embeddings are used only during graph construction for content-word bridging and do not enter the classification features.

Table~\ref{tab:efficiency_detail} summarizes the computational profile of each pipeline component.
Scikit-learn strategies train in under one second on a single CPU core.
The GAT models have 335,247--511,503 parameters (5--8~MB each), making them small by current standards.
Per-run parameter counts of 506,923--846,701 are two to three orders of magnitude smaller than the transformer-based systems that dominate the top of the accuracy leaderboard, where parameter counts range from hundreds of millions to over 100 billion.

\begin{table}[t]
  \caption{Computational profile by pipeline stage. All timings measured on a single machine (CPU only for sklearn; GPU optional for GAT).}
  \label{tab:efficiency_detail}
  \begin{tabular}{lrr}
    \toprule
    Component & Parameters & Train time \\
    \midrule
    Stanza parsing        & (pretrained) & n/a (inference only) \\
    Graph construction    & 0            & $<$1\,s per doc \\
    Feature extraction    & 0            & $<$1\,s per doc \\
    sklearn classifiers   & $<$1K        & 0.2--0.6\,s total \\
    GAT architectures     & 335K--512K   & 35--599\,s (5-fold CV) \\
    \bottomrule
  \end{tabular}
\end{table}

The HIPE-2026 Efficiency profile ranks runs by the arithmetic mean of three ranks (accuracy rank, model parameter-count rank, and deployed model-size rank), so a lower efficiency score is better~\cite{opitz_extended_2026}. For run $r$ the efficiency score is $\tfrac{1}{3}\left(\mathrm{rank}_{\mathrm{acc}}(r)+\mathrm{rank}_{\mathrm{params}}(r)+\mathrm{rank}_{\mathrm{size}}(r)\right)$.
Run~1 placed 3rd overall despite ranking only 26th on accuracy, because its small parameter budget and model size compensated.
Run~1 ranked 2nd in English and 5th in German; Run~2 ranked 2nd in French.

\begin{figure}[t]
  \centering
  \includegraphics[width=0.82\linewidth]{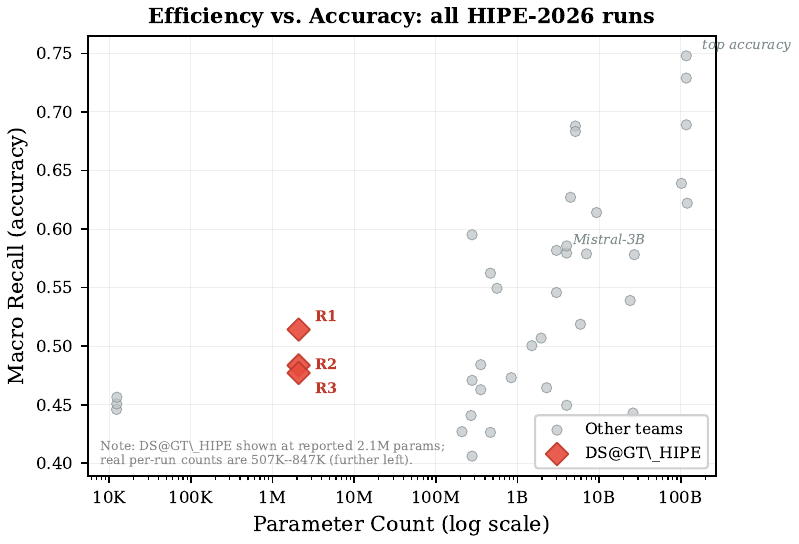}
  \caption{Accuracy versus efficiency rank for the three submitted runs (filled markers) and reference systems. Our runs cluster in the high-efficiency, mid-accuracy region of the plot.}
  \label{fig:efficiency}
\end{figure}

Figure~\ref{fig:efficiency} plots macro recall against efficiency rank for the submitted runs, illustrating the accuracy--efficiency trade-off.
The accuracy gap to the top system (0.7479 versus our best of 0.5142) is 23 percentage points.
However, the top systems rely on large language models with billions of parameters and substantially higher inference costs.
Our system demonstrates that engineered features over dependency parses can reach mid-table accuracy at a fraction of the computational cost, which is the central trade-off this work explores.
For historical newspaper processing at scale, where millions of entity pairs must be classified, this cost difference is substantial~\cite{schwartz_greenai2020,strubell_energy2019}.

\subsection{Error Analysis}
\label{sec:errors}

To characterize the failure modes of our system, we examined per-class performance on English held-out data (17 documents, 151 pairs) using the pipeline that produced the submitted runs.
Table~\ref{tab:perclass} reports precision and recall by class for both relations.

\begin{table}[t]
  \caption{Per-class precision and recall for \texttt{at} and \texttt{isAt} on English held-out data (17 documents, 151 entity pairs).}
  \label{tab:perclass}
  \begin{tabular}{llrrr}
    \toprule
    Relation & Class & Support & Precision & Recall \\
    \midrule
    \texttt{at} & FALSE    & 68 & 0.56 & 0.78 \\
    \texttt{at} & PROBABLE & 54 & 0.47 & 0.39 \\
    \texttt{at} & TRUE     & 29 & 0.42 & 0.17 \\
    \midrule
    \texttt{isAt} & FALSE  & 133 & 0.88 & 0.98 \\
    \texttt{isAt} & TRUE   & 18 & 0.25 & 0.06 \\
    \bottomrule
  \end{tabular}
\end{table}

The classifier identifies FALSE accurately for both relations but performs poorly on TRUE.
For \texttt{at}, only 5 of 29 gold TRUE pairs are correctly identified (recall 0.17); for \texttt{isAt}, only 1 of 18 (recall 0.06).
The confusion between PROBABLE and TRUE for \texttt{at} is the largest source of error, suggesting that the proximity features that separate FALSE from non-FALSE do not discriminate well between degrees of certainty.
The extreme class imbalance in \texttt{isAt} (133 FALSE versus 18 TRUE on held-out) compounds this problem, as the model learns a strong FALSE prior that it rarely overrides.

Analysis of individual false negatives reveals two dominant failure patterns.
First, OCR-broken entity surface forms prevent correct entity-to-head-word mapping.
For example, in document \texttt{sn89058133-1920-07-08}, the person mention ``Frank\textbackslash nShirley'' (with a line-break mid-name) and the location ``Cookeville'' are gold TRUE, but the system predicts PROBABLE because the broken mention disrupts the dependency parse path.
Similarly, ``Miss Williams'' paired with ``Shelby coun\textbackslash nty'' (split location) is misclassified for the same reason.
Second, cross-paragraph entity pairs with large character distances are systematically missed. ``OTTO'' paired with ``Hereford street'' at a character distance of 765 has no graph path between the entities, forcing the classifier to rely on a single noisy feature.
Of the 152 entity pairs with no graph path in the English training data, 93\% are due to graph fragmentation (the person and location are correctly tagged but fall in different connected components) rather than entity-tagging failure, and most of these disconnected pairs are structurally cross-paragraph with no shared topical content.

False positives exhibit a complementary pattern.
In short articles containing few entities, the model treats proximity as evidence of a TRUE relation.
For instance, ``Moses'' (a religious reference, not a person at a location) paired with ``Nashville, Tenn.'' is predicted TRUE because the character distance is small and the article is brief.
``Mr.\ Pain Wareing'' paired with ``D.\ C.'' is similarly a proximity overfire in a short article where the person and location co-occur without a spatial relation.
These cases suggest that the system conflates textual proximity with semantic relatedness, a limitation inherent to distance-based features that Giuliano et al.~\cite{giuliano_shallow2006} also identified in biomedical RE. Shallow features are resilient to parser noise but cannot distinguish co-occurrence from genuine relation.

\subsection{Parse Quality and the GAT--Sklearn Gap}
\label{sec:parse}

The per-language gap between GAT and sklearn performance varies substantially. On internal evaluation, the GAT provides 2.8 percentage points of lift over the best sklearn pass on English but offers no improvement on French.
We audited parse quality across languages to investigate this variation.
Table~\ref{tab:parse} reports the no-path rate (fraction of entity pairs with no graph path) and orphan entity rates (fraction of person or location nodes with degree zero in the subgraph) per language.

\begin{table}[t]
  \caption{Parse quality metrics per language, aggregated across all datasets. Higher values indicate worse parse quality.}
  \label{tab:parse}
  \begin{tabular}{lrrr}
    \toprule
    Language & No-path rate & Orphan PER & Orphan LOC \\
    \midrule
    English & 0.320 & 0.193 & 0.166 \\
    German  & 0.278 & 0.182 & 0.130 \\
    French  & 0.219 & 0.125 & 0.109 \\
    \bottomrule
  \end{tabular}
\end{table}

French has the highest parse quality, with the lowest no-path rate (22\%), the densest subgraphs, and the lowest orphan rates.
English has the worst, with roughly 1 in 3 entity pairs having no graph path and 1 in 5 person mentions having degree zero in the subgraph.
German is intermediate.

These differences explain the GAT--sklearn pattern through two compounding effects.
When parse quality is poor and training data is small (English, 307 pairs), the 15 scalar features extracted from noisy edges are themselves noisy, and the sklearn classifiers do not have enough training examples to denoise them.
The GAT, operating on the same noisy graph, integrates evidence from per-node embeddings and multi-hop attention in a way that the 15-scalar projection cannot.
When parse quality is good and training data is larger (French, 478 pairs), the scalar features are cleaner, and sklearn has enough examples to learn effective decision boundaries.
The graph's marginal benefit shrinks, and tabular models close the gap.
The submitted GAT architectures are consistent with this interpretation. For French, whose parses are cleanest, the newspaper held-out sweep selected the non-graph EndpointMLP (Table~\ref{tab:gat_sweep}), indicating that when the parse is reliable the node features alone suffice; for English, whose parses are noisiest, the graph-based PairGATStructuralSkip was selected; and for the intermediate-quality German parses PairGAT was selected. Graph message passing thus appears to help more as parse quality degrades.

\section{Conclusion}

We presented a lightweight pipeline for person--place relation extraction from multilingual historical newspapers, combining dependency-parse graphs, 15 engineered proximity features, and small classifiers with per-run parameters under 847K.

The central finding is that minimum character distance dominates the classification signal.
Feature ablation showed that adding groups beyond Group~A provides inconsistent gains and sometimes degrades performance.
A deterministic path-length rule outperformed all trained models on internal evaluation (0.549 versus 0.520 for the best sklearn pass), and the GAT branch offered no meaningful improvement over tabular classifiers.
These results align with prior work on proximity dominance in RE~\cite{jiang_features2007,zeng_position2014,alt_probing2020} and suggest that the task partly rewards surface proximity rather than semantic understanding.

Error analysis revealed complementary failure modes.
The system misses TRUE relations at distance (93\% of no-path pairs are due to graph fragmentation, not tagging failure) and hallucinates them at proximity (short articles where co-occurrence does not imply a spatial relation).
Parse quality varies across languages and explains the GAT--sklearn gap. English (no-path rate 32\%, orphan PER 19\%) benefits from the GAT's richer representation, while French (no-path 22\%, orphan PER 13\%) allows tabular models to match the graph-based approach.
Document-grouped cross-validation proved essential, revealing 25--37 percentage point inflation over naive splits~\cite{roberts_cv2017,elangovan_leakage2021}.

Several directions remain open. New bridge types, for example coreference-based or discourse-based links, could extend coverage beyond the current 69\% of reachable pairs, directly targeting the cross-paragraph failures identified in the error analysis. Reformulating \texttt{at} as a binary cascade (first FALSE versus non-FALSE, then TRUE versus PROBABLE) could address the confusion between TRUE and PROBABLE that dominates our \texttt{at} errors. A dedicated domain-adaptation path for Test~B, in place of reusing the French newspaper models unchanged, would likely recover much of the accuracy lost to register shift. Class-imbalance handling for the rare TRUE class, and calibrated thresholds tuned per language, are natural extensions given the strong FALSE prior on \texttt{isAt}. Finally, evaluation on larger corpora would test whether the lightweight, feature-driven approach continues to hold its efficiency advantage as training data grows.

\clearpage
\appendix

\section{Supplementary Results}
\label{sec:appendix}

This appendix provides detailed results that support the analyses in the main text.

\subsection{GAT Architecture Sweep}

Table~\ref{tab:gat_sweep} reports the full architecture sweep across all four GAT variants on newspaper held-out data at seed~1.
These results informed the per-language architecture selection summarized in Table~\ref{tab:gat_cv}.

\begin{table}[ht]
  \caption{GAT architecture sweep on newspaper held-out data (seed~1). Global is the mean of \texttt{at} and \texttt{isAt} macro recall. Bold marks the per-language winner.}
  \label{tab:gat_sweep}
  \begin{tabular}{llrrr}
    \toprule
    Language & Architecture & \texttt{at} & \texttt{isAt} & Global \\
    \midrule
    EN & PairGAT              & 0.378 & 0.542 & 0.460 \\
    EN & PairGATBranched       & 0.356 & 0.583 & 0.469 \\
    EN & \textbf{PairGATStructuralSkip} & \textbf{0.459} & \textbf{0.625} & \textbf{0.542} \\
    EN & EndpointMLP           & 0.333 & 0.583 & 0.458 \\
    \midrule
    FR & PairGAT              & 0.383 & 0.689 & 0.536 \\
    FR & PairGATBranched       & 0.333 & 0.595 & 0.464 \\
    FR & PairGATStructuralSkip & 0.307 & 0.662 & 0.485 \\
    FR & \textbf{EndpointMLP}  & \textbf{0.388} & \textbf{0.694} & \textbf{0.541} \\
    \midrule
    DE & \textbf{PairGAT}     & \textbf{0.411} & \textbf{0.629} & \textbf{0.520} \\
    DE & PairGATBranched       & 0.389 & 0.564 & 0.477 \\
    DE & PairGATStructuralSkip & 0.371 & 0.543 & 0.457 \\
    DE & EndpointMLP           & 0.268 & 0.447 & 0.357 \\
    \bottomrule
  \end{tabular}
\end{table}

\subsection{Sklearn Pass Comparison on Held-Out Data}

Table~\ref{tab:passes} compares the three sklearn pass designs on newspaper held-out data (7 documents per language).
Pass~A trains a fixed classifier pair (RandomForest for \texttt{at}, ExtraTrees for \texttt{isAt}) per language.
Pass~B selects the per-language CV winner for each relation independently.
Pass~C trains on multilingual-concatenated data from all three languages.

\begin{table*}[ht]
  \caption{Sklearn held-out pass comparison (7 docs/lang). Bold marks the best per-language global score. Mean global is the average across all three languages.}
  \label{tab:passes}
  \begin{tabular}{lrrrrrrrrrrr}
    \toprule
    & \multicolumn{3}{c}{English} & \multicolumn{3}{c}{French} & \multicolumn{3}{c}{German} & Mean \\
    \cmidrule(lr){2-4} \cmidrule(lr){5-7} \cmidrule(lr){8-10}
    Pass & \texttt{at} & \texttt{isAt} & Global & \texttt{at} & \texttt{isAt} & Global & \texttt{at} & \texttt{isAt} & Global & Global \\
    \midrule
    A & 0.421 & 0.625 & \textbf{0.523} & 0.383 & 0.732 & 0.558 & 0.331 & 0.593 & 0.462 & 0.514 \\
    B & 0.495 & 0.487 & 0.491 & 0.406 & 0.759 & \textbf{0.583} & 0.331 & 0.564 & 0.448 & 0.507 \\
    C & 0.385 & 0.583 & 0.484 & 0.395 & 0.716 & 0.555 & 0.403 & 0.636 & \textbf{0.519} & \textbf{0.520} \\
    \bottomrule
  \end{tabular}
\end{table*}

\subsection{Feature Importance Rankings}

Table~\ref{tab:importance} reports RandomForest feature importance scores (Gini impurity-based) for the top-ranked features on English newspaper data.
The four highest-ranked features for both relations are proximity-based, corroborating the ablation analysis in Section~\ref{sec:ablation}.

\begin{table}[ht]
  \caption{Top 5 RandomForest feature importance scores for \texttt{at} and \texttt{isAt} on English newspaper data.}
  \label{tab:importance}
  \begin{tabular}{rlr}
    \toprule
    Rank & Feature & Importance \\
    \midrule
    \multicolumn{3}{l}{\textit{at}} \\
    1 & l\_position\_norm     & 0.151 \\
    2 & min\_char\_dist       & 0.150 \\
    3 & p\_position\_norm     & 0.124 \\
    4 & sent\_dist            & 0.089 \\
    5 & l\_location\_align    & 0.088 \\
    \midrule
    \multicolumn{3}{l}{\textit{isAt}} \\
    1 & l\_position\_norm     & 0.177 \\
    2 & min\_char\_dist       & 0.162 \\
    3 & p\_position\_norm     & 0.119 \\
    4 & sent\_dist            & 0.112 \\
    5 & min\_char\_dist\_via\_pron & 0.051 \\
    \bottomrule
  \end{tabular}
\end{table}

\clearpage
\section*{Declaration on Generative AI}
During the preparation of this work, the author used generative AI tools for grammar and spelling checking, for aggregating and formatting outputs generated by the author's own analysis scripts, and for code generation, correction, and review. All research ideas, concepts, methods, experimental design, and interpretations are the author's own. After using these tools, the author reviewed and edited all content as needed and takes full responsibility for the publication's content.

\bibliography{references}

\end{document}